%% file: acl_latex.tex
\documentclass[11pt]{article}

\usepackage[final]{acl}

\usepackage{times}
\usepackage{latexsym}

\usepackage[T1]{fontenc}

\usepackage[utf8]{inputenc}

\usepackage{microtype}

\usepackage{inconsolata}

\usepackage{graphicx}

\usepackage{subcaption}
\usepackage{multirow}
\usepackage{booktabs}
\usepackage{enumitem}
\usepackage{amsmath}
\usepackage{url}
\title{The RAT: A Unified Bayesian Model for RAG Evaluation}

\author{
\textbf{Pius von D\"{a}niken}\thanks{Equal contribution} \quad
\textbf{Felix Matthias Saaro}\footnotemark[1]
\\
\textbf{Mark Cieliebak} \quad
\textbf{Jan Milan Deriu}
\\
Centre for Artificial Intelligence\\
ZHAW School of Engineering\\
\texttt{\{vode,saaf,ciel,deri\}@zhaw.ch}
}

\begin{document}
\maketitle
\begin{abstract}
Evaluating Retrieval-Augmented Generation (RAG) systems requires assessing not only end-to-end correctness but also how individual components interact and how errors propagate through the pipeline. We introduce a Bayesian evaluation framework that jointly models retrieval success, abstention behavior, and answer correctness, factorized according to the pipeline's information flow. The model distinguishes \emph{task success}. whether the user received a correct answer, from \emph{generator success}, whether the generator behaved appropriately given the retrieval outcome. We apply the framework to 27 RAG configurations across three datasets, three retrievers, and three generators, and show that the conditional decomposition reveals substantial behavioral differences between systems that appear equivalent under marginal metrics. We further analyze the annotation allocation problem, demonstrating that retrieval-success annotations are more informative than task-success annotations for estimating policy adherence, and provide an information-theoretic explanation for this asymmetry. Finally, we extend the model to incorporate LLM-as-a-judge annotations as calibrated noisy observations, enabling practitioners to combine limited human judgments with cheaper automated assessments within a unified probabilistic model.
\end{abstract}

\section{The Introduction}


\begin{figure}
    \centering
    \includegraphics[width=0.85\linewidth]{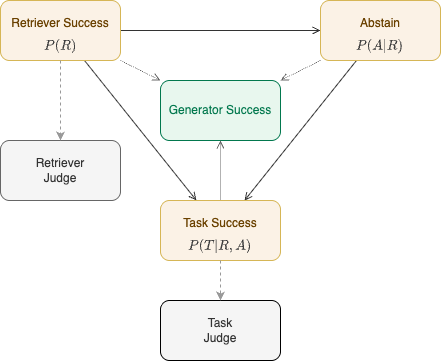}
    \caption{The dependency structure of the evaluation variables: generator success depends on retrieval success (which governs whether the generator should abstain or answer), abstention behavior, and task success.}
    \label{fig:overview}
\end{figure}

Evaluating Retrieval-Augmented Generation (RAG) systems is challenging and often tedious. It is not sufficient to evaluate only the system's end-to-end behavior; each component must be evaluated both in isolation and within the pipeline. Since RAG systems are built in a pipelined approach, errors tend to propagate through the pipeline. If retrieval fails, the generator has no basis for producing a correct answer and should abstain from answering the question. Most benchmarks developed for RAG evaluation consider each component in isolation~\cite{es2024ragas, saad-falcon-2024-ares,rau-etal-2024-bergen}, or evaluate generators using adversarial retrieval results~\cite{wang2024feb4rag}. However, a holistic view of the evaluation that accounts for the dependencies between components is still missing.

A key shortcoming of end-to-end evaluation is that it conflates two distinct notions of success. \emph{Task success} measures whether the user received a correct answer, whereas \emph{generator success} measures whether the generator behaved appropriately given the retrieval outcome, i.e., answering when retrieval succeeded and abstaining when it failed. These two quantities can diverge substantially: a generator that never abstains may achieve reasonable task success while systematically violating the desired policy, and two systems with identical task success can exhibit very different behaviors under retrieval failure. To make such distinctions explicit, we need a model that captures the dependencies between retrieval, abstention, and answer correctness.

We propose a Bayesian evaluation framework that models these dependencies explicitly. Figure~\ref{fig:overview} illustrates the dependency structure, which shows how these variables interact: retrieval success governs both the abstention decision and the conditions under which task success is meaningful, and generator success is a deterministic function of all three. Casting this as a Bayesian model allows us to propagate uncertainty through the full dependency structure and to handle partially observed data by marginalizing over unobserved variables, which is a practical advantage when some annotations are expensive while others are cheap. The framework is further extended by incorporating LLM-as-a-judge annotations alongside human gold-standard labels, using a calibration model that treats automated judgments as noisy observations of the underlying ground-truth variables~\cite{von-daniken-etal-2022-effectiveness,von2024improving}. This allows practitioners to combine small amounts of expensive human annotation with larger volumes of automated judgments in a principled way. We make the following contributions~\footnote{Our code can be found at:~\url{https://github.com/vodezhaw/rat}.}:
\begin{itemize}[nosep,leftmargin=*]
    \item We introduce a Bayesian evaluation model for RAG systems that factorizes the joint distribution over retrieval success, abstention, and task success according to the pipeline's information flow, and derives generator success as a deterministic variable over these random variables (Section~\ref{sec:model}).
    \item We apply the model to 27 RAG configurations (3~retrievers $\times$ 3~generators $\times$ 3~datasets) and show that the conditional decomposition reveals behavioral differences that marginal metrics conceal, in particular, systems with near-identical task success can differ sharply in policy adherence.
    \item We analyze the annotation allocation problem: given a fixed budget, we derive and empirically validate which combination of retrieval and task-success annotations minimizes estimation error, and provide an information-theoretic explanation for the observed asymmetry.
    \item We extend the model to incorporate calibrated LLM-as-a-judge annotations as noisy observations, enabling practitioners to combine human and automated judgments within the same probabilistic framework.
\end{itemize}

\section{The Related Work}
Retrieval-Augmented Generation (RAG) is the process of incorporating the output of an information retrieval (IR) system into the context of a large language model (LLM), so that the LLM's answer is informed by relevant documents or passages~\cite{lewis2020orig_rag,borgeaud2022rag}. The process is composed of two components: retrieval and generation. The user's question is transformed into a query for an IR engine, which retrieves a set of relevant documents or passages, which are then given to the LLM to generate an answer to the user's query. This setting, while straightforward to describe, is highly challenging to evaluate. In general, the two components are evaluated separately. For an extensive overview of RAG evaluation, we refer the reader to~\cite{hao2025ragevalsurvey}. 

\noindent\textbf{Retrieval Evaluation.} There are two main settings: either there is a gold standard available, that is, for a set of queries, all relevant documents, or there is no gold standard available. In the case of an existing gold standard, which is highly costly to create, one applies the standard scores used in the IR literature~\cite{tang2024multihop}, such as Mean Reciprocal Rank, or Mean Average Precision~\cite{schutze2008introduction}. In cases where no gold standard is available, two approaches are commonly used: either silver-standard generation or an unreferenced LLM-as-a-judge~\cite{zheng2023llmasajudge}. The generation of silver standards consists of an inverted process by providing an LLM with a document or passage, and letting the LLM generate questions~\cite{es2024ragas}. Then the provided document serves as the relevant document (while disregarding other potentially relevant documents). For LLM-as-a-judge, one provides the LLM with the result of the retrieval and the user question, and asks to rate whether the question can be answered with the provided retrieval~\cite{saad-falcon-2024-ares}. 

\noindent\textbf{Generator Evaluation.} The evaluation of the generator often entangles two concepts: the end-to-end success, and the evaluation of the generator under different retrieval settings. For instance, \citet{wang2024feb4rag} evaluates the behavior of the generator under different adversarial retrieval settings. \citet{chen2024llmsinrag} also highlights the importance of evaluating the impact of the retrieval on the generation in isolation. They also state the need to handle cases where the retrieval does not find relevant sources in the generator by allowing abstention from answering, and, in turn, evaluating the LLM's behavior in these cases. 

\noindent\textbf{Pipeline Evaluation.} \citet{rau-etal-2024-bergen} introduces a retrieval benchmarking library that focuses on evaluating pipeline components. ARES~\cite{saad-falcon-2024-ares} introduces a LLM-as-a-judge-based evaluation framework to evaluate each component of the pipeline separately using various dimensions. RAGAs~\cite{es2024ragas} introduces a reference-free framework for evaluating multiple dimensions of RAG pipelines without relying on ground-truth human annotations, covering context relevance, faithfulness, and answer relevance. Both RAGAs and ARES follow the RAG Triad framing popularized by \emph{TruLens}~\citep{trulens-rag-triad}. CRUD-RAG~\cite{yuanjie2025rageval} is a benchmark oriented towards real-world scenarios rather than the QA tasks popular in academic settings. They include evaluation of multiple components, such as chunk size selection and reranking. While these frameworks evaluate multiple dimensions, they treat each dimension independently and do not model the statistical dependencies between retrieval and generation outcomes.

\noindent\textbf{Our Work} builds on these efforts and integrates them into a single probabilistic framework that jointly models the dependencies between retrieval, abstention, and answer correctness. This unified view enables distinguishing policy adherence from end-task success, propagating uncertainty across the pipeline, reasoning about annotation allocation, and combining human and automated judgments via calibration.

\section{The Model}\label{sec:model}

Evaluating a RAG system only by whether its final output matches a reference answer does not tell the full story. End-task correctness quantifies the system's overall output, but it does not distinguish among the generator's underlying behaviors. In particular, the common intuition that better retrieval should lead to better answers is only valid if the generator can use the retrieved information appropriately. Likewise, abstention can be desirable when retrieval fails, even though it may reduce overall answer rates and, in turn, affect aggregate performance metrics.

\noindent\textbf{The Variables.} To make these distinctions explicit, we model RAG behavior using four binary variables. Let $R$ denote retrieval success, where $R=1$ means that the retrieved context contains all information required to answer the question. Let $A$ denote abstention, where $A=1$ means that the generator declines to answer by producing an explicit "I don't know." response. Let $T$ denote task success, where $T=1$ means that the final answer is correct. Finally, let $G$ denote generator success, or policy adherence, where $G=1$ if the generator behaves as desired. We use binary variables as a deliberate simplification.  

\noindent\textbf{The Conditionals.} Studying only the marginals of these variables is insufficient to characterize the system's behavior. For example, a system with slightly higher task success may still be less desirable if it achieves that gain by answering questions without appropriate support, rather than abstaining appropriately. To understand such trade-offs, we study conditional probabilities that separate retrieval quality, abstention behavior, and answer correctness.

We therefore propose a factorization of the joint distribution that follows the flow of information through the RAG pipeline:
\[
P(R, A, T) = P(R)\,P(A \mid R)\,P(T \mid A, R).
\]
In our data (see Section~\ref{sec:data}), task success is only possible for non-abstained responses, so effectively $T=0$ whenever $A=1$. This yields conditional probabilities with direct behavioral interpretations: $P(R)$ captures retrieval quality, $P(A \mid R)$ captures how abstention depends on retrieval success or failure, and $P(T \mid A=0, R)$ captures answer correctness conditional on both the retrieval state and the decision to answer.

\noindent\textbf{The Generator Success.} Finally, we define generator success $G$ deterministically from $(R,A,T)$ to encode the desired policy:
$G = \bigl((R=0) \land (A=1)\bigr)\ \lor\ \bigl((R=1) \land (A=0) \land (T=1)\bigr).$
That is, the generator is successful if it abstains when retrieval fails, or if it answers correctly when retrieval succeeds. This construction makes it possible to distinguish correct behavior from correct output and to analyze how different RAG configurations trade off retrieval, abstention, and answer quality.

\noindent\textbf{The Unified Model.} We cast this model in a Bayesian framework. This provides two practical advantages for our setting. First, it propagates uncertainty throughout the full model, allowing us to quantify uncertainty not only for the primitive conditional probabilities but also for derived quantities such as policy adherence. Second, it naturally accommodates partially observed data by marginalizing over unobserved variables, which is particularly useful when some annotations are expensive while others are cheap.

Under the factorization above, the model is parameterized by five probabilities:
$\theta_R = P(R=1)$, $\theta_{A^-} = P(A=1 \mid R=0)$, $\theta_{A^+} = P(A=1 \mid R=1)$,
$\theta_{T^-} = P(T=1 \mid R=0, A=0)$, and $\theta_{T^+} = P(T=1 \mid R=1, A=0)$.
These correspond directly to interpretable aspects of RAG behavior: retrieval success, abstention on retrieval failure, abstention despite successful retrieval, unsupported-answer success, and supported-answer success.

\noindent\textbf{The Automated Judge Extension.} To add the LLM-as-a-judge to the model, we extend the factorization by an additional term to represent the binary $R_J$ and $T_J$ variables: $P(R, A, T, R_J, T_J) = P(R)P(A|R)P(T|R, A)P(R_J, T_J | R, A, T)$. This adds 18 degrees of freedom to the model, corresponding to the 6 feasible $(R, A, T)$ states and a 4-category conditional distribution for each state. The resulting judgment model is more complex than simpler factorizations such as $P(R_J | R)P(T_J|T)$, but avoids imposing conditional independence assumptions that were not supported by preliminary analyses.

We place independent uniform priors on all five basic model parameters and a Dirichlet prior on the conditional probability of automated judge observations. The resulting posterior can then be queried for both the primitive parameters and any deterministic function of them, including model-implied marginals and the derived policy-adherence quantity. We implement the model in Stan~\citep{stan} and perform posterior inference using Hamiltonian Monte Carlo with the No-U-Turn Sampler (NUTS)~\citep{NUTS}. For each experiment, we run five chains with 2{,}000 warmup iterations and collect 10{,}000 posterior samples per chain.

\section{The Retrievers, The Generators, and The Tasks}\label{sec:data}

\noindent\textbf{The Data.} This work employs the KILT benchmark \cite{petroni-etal-2021-kilt}, a unified library for knowledge intensive languag tasks comprising 11 datasets across 5 task categories, including fact-checking and open-domain question answering. All datasets are grounded in a shared, pre-processed Wikipedia snapshot, thereby ensuring consistent evaluation conditions and facilitating interoperability across tasks with minimal preprocessing overhead.

\begin{itemize}[noitemsep]
    \item \textbf{FEVER (FEV)} \cite{Thorne18Fever} is a fact-checking dataset consisting of approximately 126{,}000 human-annotated claims, each assigned one of three labels: \textit{Supported}, \textit{Refuted}, or \textit{NotEnoughInfo}. The majority of queries have 1 relevant paragraph.
    \item \textbf{HotpotQA (HQA)} \cite{yang2018hotpotqa} is an open-domain question answering dataset comprising approximately 113{,}000 questions that necessitate multi-hop reasoning across multiple Wikipedia articles. The question-answer pairs were collected using crowd workers who were shown pairs of Wikipedia paragraphs and asked to write multi-hop questions along with their answer. All queries have 2 relevant paragraphs.
    \item \textbf{Natural Questions (NQ)} \cite{kwiatkowski-etal-2019-nq} consists of approximately 92{,}000 naturally occurring queries submitted to the Google search engine, each paired with a long and short answer annotation curated by crowd workers. In this work, only the short answer annotations are used. All queries have exactly 1 relevant paragraph.
\end{itemize}

\noindent For each task, a subset of 10{,}000 queries was randomly selected. A single shared knowledge base of 1{,}720{,}160 paragraphs was subsequently constructed from all documents relevant to the selected queries across all tasks\footnote{This knowledge base represents a more controlled retrieval environment than a fully open-domain corpus. Consequently, the absolute retrieval-success rates (see Section~\ref{sec:main-res}) may be interpreted as optimistic relative to fully open-domain deployment.}.


\noindent\textbf{The Retrieval Strategies.}
Three strategies were evaluated:
\begin{itemize}[noitemsep,leftmargin=*]
    \item  \textbf{Sparse (S)}, a term-based retrieval approach employing the BM25 ranking function based on lexical matching.
    \item \textbf{Dense (D)}, semantic retrieval approach leveraging multi-task text embeddings  \cite{sturua2024jinaembeddingsv3multilingualembeddingstask}, indexed using a Hierarchical Navigable Small World (HNSW) graph via the FAISS library \cite{douze2024faiss} for efficient approximate nearest-neighbour search.
    \item \textbf{Hybrid (H)}, a combination of sparse and dense retrieval, with result fusion performed using Reciprocal Rank Fusion (RRF) \cite{rrf-10.1145/1571941.1572114}.
\end{itemize}

\noindent\textbf{The Generators.}
Three large language models were employed as generators:
\begin{itemize}[noitemsep,leftmargin=*]
    \item \textbf{Apertus~8B} \cite{swissai2025apertus}, a fully open-source, decoder-only transformer model with 8 billion parameters, developed with an emphasis on full transparency of both model weights and training data.
    \item  \textbf{Gemma3~12B }\cite{gemmateam2025gemma3technicalreport}, a lightweight open multimodal model developed by Google, comprising 12 billion parameters and trained on 12 trillion tokens spanning more than 140 languages.
    \item \textbf{Qwen3.5~9B} \cite{qwen3.5}, a multimodal model developed by Alibaba Cloud with 9 billion parameters, supporting multilingual inference across 201 languages. Although the model includes a dedicated reasoning mode, this functionality was not employed in the present work
\end{itemize}

\section{The Experiments}\label{sec:main-res}




\begin{table}[t!]
\centering
\resizebox{\columnwidth}{!}{%
\input{tables/naive_all_marginal_point_estimates}
}
\caption{Marginal probabilities across all 27 RAG configurations. $P(R{=}1)$ is shared across generators within each dataset--retriever pair. Highest per-dataset values in \textbf{bold}, lowest in \textit{italics}.}
\label{tab:generator_retrieval_results}
\end{table}

\subsection{The Marginals}

Table~\ref{tab:generator_retrieval_results} reports the marginal probabilities across all 27 configurations. We count retrieval as successful only when \emph{all} relevant documents are present in the retrieved context\footnote{This binary definition is a modeling simplification (see Section~\ref{sec:model}). We discuss partial retrieval in Appendix~\ref{sec:partial-success-conditionals}.}. Hybrid retrieval achieves the highest retrieval success on \textit{FEV} and \textit{NQ}, whereas sparse retrieval performs best on \textit{HQA}. Retrieval success is substantially higher on \textit{FEV} than on the other two datasets because each sample has exactly one relevant document, making the retrieval task considerably easier. Turning to the generator-level metrics, \textit{Apertus} exhibits consistently low abstention rates across all datasets, with a maximum of 0.135 on \textit{FEV} with dense retrieval. In contrast, \textit{Gemma3} and \textit{Qwen3.5} abstain substantially more often, reaching rates as high as 0.489 for \textit{Qwen3.5} on \textit{HQA} with dense retrieval.



Task success is highest on \textit{FEV}, where each question has only two answer options, and lower on \textit{HQA} and \textit{NQ}, which require an exact match to an open-form reference answer. Across datasets, the three generation models achieve broadly similar task success rates. Policy adherence, however, differs much more strongly across models: \textit{Qwen3.5} achieves the highest policy adherence on all datasets, whereas \textit{Apertus} consistently lags behind.


\begin{table}[t!]
    \centering
    \small
    \input{tables/conditional_point_estimates_compressed}
    \caption{
    Model-implied conditional probabilities $\theta_{A^-}$, $\theta_{A^+}$, $\theta_{T^-}$, and $\theta_{T^+}$ across all 27 RAG configurations, reported as posterior mean estimates. We show the highest value for each dataset in {\bf{bold}} and the lowest in \it{italics}.
    }
    \label{tab:rag-conditionals}
\end{table}

\subsection{The Conditionals}
We fit the model introduced in Section~\ref{sec:model} separately to the data from each configuration. The model-implied marginals closely match the corresponding count-based marginals up to Monte Carlo error. Table~\ref{tab:rag-conditionals} therefore focuses on the conditional probabilities, which provide a more interpretable decomposition of system behavior.

A first pattern is that \textit{Gemma3} and \textit{Qwen3.5} distinguish between retrieval success and failure much more clearly than \textit{Apertus}. Across datasets, both models assign substantially higher abstention probabilities when retrieval fails than when retrieval succeeds. For example, on \textit{HQA} with hybrid retrieval, \textit{Apertus} has $P(A{=}1 \mid R{=}0)=0.039$ and $P(A{=}1 \mid R{=}1)=0.005$, whereas \textit{Gemma3} reaches $0.270$ and $0.021$, and \textit{Qwen3.5} $0.463$ and $0.036$. Thus, under the same retrieval conditions, the models exhibit markedly different abstention policies. At the same time, abstention conditional on retrieval success remains low across all models, indicating that unnecessary abstention is uncommon.

A second pattern is that retrieval success consistently improves answer success among non-abstained responses, but the magnitude of this improvement depends strongly on the dataset. On \textit{FEV}, unsupported task success remains high even when retrieval fails, reflecting the relative ease of the task and the fact that each question has only two answer options. By contrast, on \textit{HQA} and \textit{NQ}, where retrieval success requires recovering all necessary documents and answers must exactly match an open-form reference, the gap between $P(T{=}1 \mid R{=}0, A{=}0)$ and $P(T{=}1 \mid R{=}1, A{=}0)$ is much larger.

The conditional decomposition also clarifies why similar marginal task success can hide substantial behavioral differences. On \textit{NQ} with dense retrieval, the three generators achieve nearly identical task success rates (\textit{Apertus}: $0.239$, \textit{Gemma3}: $0.242$, \textit{Qwen3.5}: $0.243$), yet their policy adherence differs sharply (\textit{Apertus}: $0.164$, \textit{Gemma3}: $0.414$, \textit{Qwen3.5}: $0.496$). In other words, comparable end-task accuracy does not imply comparable behavior under retrieval failure. The conditional probabilities reveal that these differences are largely driven by the abstention policy rather than by answer accuracy alone.

Finally, improved retrieval does not automatically translate into proportional gains in either task success or policy adherence. Better retrieval increases the opportunity to answer correctly, but the realized benefit depends on whether the generator both recognizes retrieval failure and uses retrieved support effectively when it is available.

\subsection{The Sample Allocation Problem}
Here, we investigate a setting closer to what one might encounter in real-world applications, where annotation scarcity is common. Thus, we assume access to 100 annotations (\emph{Base} samples) for retrieval and task success, and that abstention is always observed (via simple string matching). Then, we assume that we are given an additional budget for more annotations (\emph{Add.} samples). The question is where to allocate the additional samples: to measuring retrieval success, task success, or a mix of the two. We apply the model to 100 Base samples with full annotations for retrieval and task success, assuming that abstention is always observed via simple string matching. Given an additional annotation budget of $\text{\emph{Add}}. \in {60, 100, 200, 500}$ samples, we investigate five allocation strategies: (1) all additional samples receive full annotations (i.e., both task-success and retrieval-success annotations), (2) half receive full annotations and half only retrieval-success annotations, (3) half receive full and half only task-success annotations, (4) all additional samples receive only retrieval-success annotations, and (5) all additional samples receive only task-success annotations. To ensure robust estimates, we subsample 500 times from the 10,000 available samples and report the Mean Absolute Error (MAE) relative to the full-data point estimate, as well as the 95\% credible interval width. We run experiments on the HQA dataset for Apertus and Qwen using the Hybrid retriever.

\begin{figure*}
    \centering
    \includegraphics[width=1\linewidth]{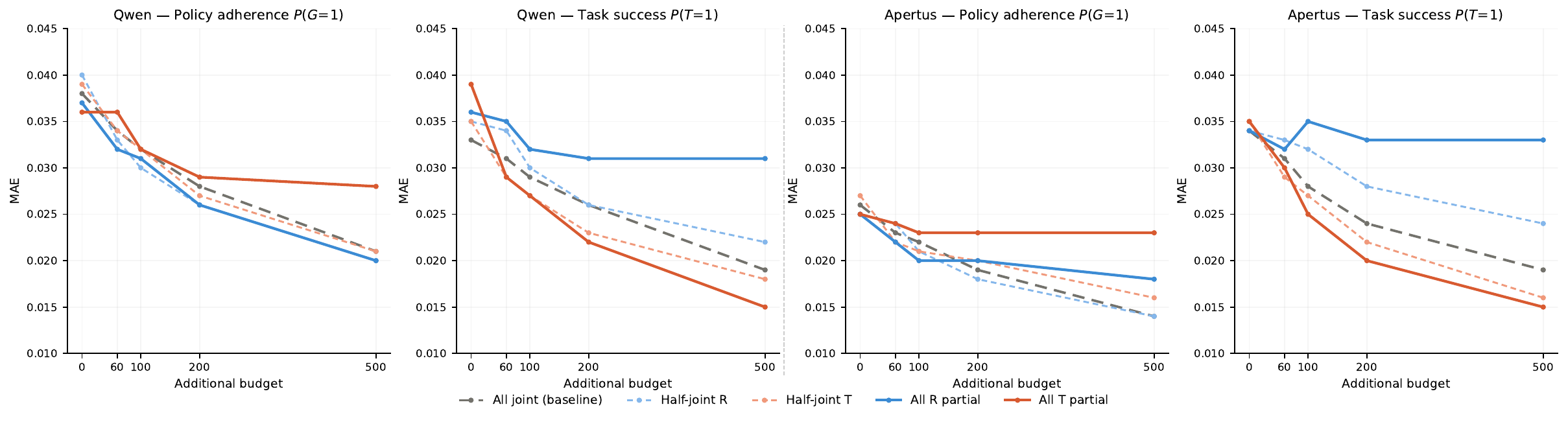}
    \caption{MAE for policy adherence $P(G{=}1)$ and task success $P(T{=}1)$ under five annotation allocation strategies, for Qwen (left) and Apertus (right). All configurations start with 100 fully annotated base samples. Abstention is always observed. Results on HotpotQA with Hybrid retrieval, averaged over 500 subsamples.}
    \label{fig:allocation}
\end{figure*}

\noindent\textbf{The Observation.} Figure~\ref{fig:allocation} reveals a consistent asymmetry across both models: retrieval-focused strategies reduce estimation error more effectively for policy adherence (G), while task-focused strategies are more effective for task success (T). For $P(G{=}1)$, the all-R strategy matches or approaches the all-joint baseline for Qwen, whereas all-T shows markedly slower improvement and plateaus at a higher MAE. This pattern holds for Apertus, though the gap between all-R and all-joint widens at larger budgets. For $P(T{=}1)$, the pattern reverses: all-T closely tracks the all-joint baseline for both models, while all-R provides negligible improvement; its MAE and CI width remain nearly flat regardless of budget, particularly for Apertus. The half-joint strategies consistently fall between the corresponding extremes, with half-joint-R closer to all-joint for $P(G{=}1)$ and half-joint-T closer to all-joint for $P(T{=}1)$. Notably, the all-joint strategy never underperforms the best partial strategy by a large margin, making it a robust default when the estimation target is not known in advance.

\noindent\textbf{The Explanation.} To understand this asymmetry, we analyze the information gain of each strategy for estimating $P(G{=}1)$ (the case for $P(T{=}1)$ is straightforward, since task-success annotations directly observe $T$). Since the abstention $A$ is always observed, we measure the conditional information gain $I(G; \cdot \mid A)$ of additionally observing $R$, $T$, or both.
The key insight follows from the definition of $G$: of the four $(R, A)$ cells, three resolve $G$ deterministically, only $(R{=}1, A{=}0)$ requires knowing $T$ (see Table~\ref{tab:ra_cells_main}). Thus, observing $R$ resolves $G$ in three of four cases, while observing $T$ resolves $G$ only in one of three non-zero $(A, T)$ cells. This structural asymmetry explains why retrieval annotations are consistently more informative for $G$ than task annotations.
Table~\ref{tab:info_gain_G_main} quantifies this using the conditional probabilities from Table~\ref{tab:rag-conditionals} (full derivations in Appendix~\ref{sec:info_gain}). For both models, the All-R strategy captures the majority of the information provided by the All-Joint strategy ($65.6\%$ for Qwen, $58.0\%$ for Apertus), whereas All-T captures far less ($29.3\%$ for Qwen, $40.8\%$ for Apertus). The difference between models reflects their abstention behavior: Qwen's strong retrieval, abstention dependency ($P(A{=}1 \mid R{=}0) = 0.463$ vs.\ $P(A{=}1 \mid R{=}1) = 0.036$) means that $R$ observations yield highly variable $G$ labels, whereas Apertus rarely abstains regardless of retrieval ($P(A{=}1 \mid R{=}0) = 0.039$), reducing the discriminative value of $R$.
The information-theoretic ranking correctly predicts the empirical ordering of partial strategies. The only discrepancy is that for Qwen, All-R slightly outperforms All-Joint at large budgets, despite lower per-sample information gain. This is because the 100 base joint samples already constrain $\theta_{T^+} = P(T{=}1 \mid R{=}1, A{=}0)$ in the single ambiguous cell, after which additional joint annotations provide diminishing returns (see Appendix~\ref{sec:info_gain} for details).


\begin{table}[t]
\centering
\small
\begin{subtable}[t]{\columnwidth}
\centering
\begin{tabular}{ccccc}
\toprule
$R$ & $A$ & $G$ & Qwen & Apertus \\
\midrule
0 & 1 & $1$ & 0.367 & 0.031 \\
0 & 0 & $0$ & 0.426 & 0.762 \\
1 & 1 & $0$ & 0.007 & 0.001 \\
1 & 0 & $T$ & 0.200 & 0.206 \\
\bottomrule
\end{tabular}
\caption{Joint probabilities $P(R,A)$ with the resulting value of $G$. In three of four cells, $G$ is determined by $(R,A)$ alone; only $(R{=}1, A{=}0)$ requires $T$.}
\label{tab:ra_cells_main}
\end{subtable}

\vspace{0.5em}

\begin{subtable}[t]{\columnwidth}
\centering
\begin{tabular}{lcc}
\toprule
Strategy & Qwen & Apertus \\
\midrule
All-Joint & 0.526 & 0.490 \\
Half-Joint-R & 0.436 & 0.387 \\
All-R & 0.345 & 0.284 \\
Half-Joint-T & 0.340 & 0.345 \\
All-T & 0.154 & 0.200 \\
\bottomrule
\end{tabular}
\caption{Information gain $I(G; \cdot \mid A)$ per sample for each annotation strategy. Higher values indicate more informative observations for estimating $P(G{=}1)$.}
\label{tab:info_gain_G_main}
\end{subtable}

\caption{Information-theoretic analysis of annotation strategies on HotpotQA with Hybrid retrieval.}
\label{tab:info_gain_combined}
\end{table}

\subsection{The Automated Judge}
Since human annotations are highly time and cost-intensive, it has become common practice to use LLM-as-a-judge to automate parts of the evaluation. We investigate the impact of using automated judgments on the evaluation pipeline. The difficulty stems from the need to calibrate automated judgments to match human judgments~\cite{von-daniken-etal-2022-effectiveness}, which introduces uncertainty that depends on the automated judge's performance. 

\noindent\textbf{The Judge.} We use gpt-4o-mini \cite{openai2024gpt4ocard} as our judge for both the task and retrieval success rates following the evaluation framework proposed by \cite{saad-falcon-2024-ares} for context and answer relevance. We compare the retrieval success rates according to the judge $P(R_J=1), P(T_J=1)$ to those according to humans $P(R=1), P(T=1)$. The judge's performance is measured in terms of true-and-false positive rates ($TPR = (R_J=1 \mid R=1)$, $FPR = P(R_J=1 \mid R=0)$). An analogous procedure is applied to task success, comparing $T_J$ with $T$ to derive the TPR and FPR.
Tables~\ref{tab:judge_retrieval} and~\ref{tab:judge_task} report calibration results for the retrieval and task judges on HotpotQA, showing that while both judges achieve high sensitivity, they exhibit substantial FPR, leading to overestimated success rates $P(R_J)$ and $P(T_J)$.


\begin{table}[t]
\centering
\small
\begin{subtable}[t]{\columnwidth}
\centering
\begin{tabular}{llcccc}
\toprule
DS & Ret. & TPR & FPR & $P(R)$ & $P(R_J)$ \\
\midrule
HQA & D & 0.77 & 0.17 & 0.15 & 0.26 \\
HQA & H & 0.77 & 0.19 & 0.21 & 0.31 \\
HQA & S & 0.77 & 0.18 & 0.22 & 0.31 \\
\bottomrule
\end{tabular}
\caption{Retrieval judge.}
\label{tab:judge_retrieval}
\end{subtable}

\vspace{0.5em}

\begin{subtable}[t]{\columnwidth}
\centering
\begin{tabular}{llcccc}
\toprule
DS & Gen. & TPR & FPR & $P(T)$ & $P(T_J)$ \\
\midrule
HQA & Qwn & 0.94 & 0.32 & 0.21 & 0.52 \\
HQA & Apt & 0.88 & 0.43 & 0.21 & 0.57 \\
HQA & Gem & 0.93 & 0.38 & 0.21 & 0.55 \\
\bottomrule
\end{tabular}
\caption{Task judge.}
\label{tab:judge_task}
\end{subtable}

\caption{LLM-as-a-judge calibration on HotpotQA. TPR and FPR denote the judge's true and false positive rates, e.g., $\text{TPR} = P(R_J{=}1 \mid R{=}1)$ and $\text{FPR} = P(R_J{=}1 \mid R{=}0)$ for retrieval. The judge exhibits high sensitivity but a substantial false-positive rate, leading to an overestimation of success rates.}
\label{tab:judge_calibration}
\end{table}

\noindent\textbf{The Judge's Impact.} We investigate whether supplementing the 200 human-annotated base samples with automated judge annotations improves estimation quality. We add $n_{\text{add}} \in \{0, 500, 5000\}$ judge-annotated samples to the base set, treating $R_J$ and $T_J$ as noisy observations of $R$ and $T$ with the calibrated TPR and FPR from Table~\ref{tab:judge_calibration}. Table~\ref{tab:judge-results} reports results for Apertus on HotpotQA with Hybrid retrieval. Adding automated judgments yields only marginal improvements: the CI width for $P(G{=}1)$ decreases from $0.094$ to $0.080$ even with 5{,}000 additional samples, while MAE remains essentially unchanged. The pattern for $P(T{=}1)$ is similar. This is consistent with the judge's high false-positive rates (Table~\ref{tab:judge_calibration}): when the FPR is substantial, each automated annotation carries limited information, and large volumes of noisy labels cannot substitute for even modest amounts of human annotation. The result highlights that the value of LLM-as-a-judge annotations depends critically on calibration quality. While our empirical findings are specific to the judges and calibration setup considered here, they illustrate that a poorly calibrated judge may add annotation volume without meaningfully reducing uncertainty. This is consistent with the findings of prior literature~\cite{von-daniken-etal-2022-effectiveness}.

\begin{table}
    \centering
    \small
    \input{tables/llm_judge_metrics_table}
    \caption{Mean absolute error and 95\% credible interval width for P(G=1) and P(T=1) based on 200 fully annotated observations and additional observations where we observe $R_J$ instead of R and $T_J$ instead of T.}\label{tab:judge-results}
\end{table}




\section{The Conclusion}

We presented a Bayesian evaluation framework for RAG systems that factorizes the joint distribution over retrieval success, abstention, and task success according to the pipeline's information flow. The key distinction is between task success, i.e., whether the user received a correct answer, and generator success, i.e., whether the generator behaved appropriately given the retrieval outcome. Applying the model to 27 configurations, we showed that systems with near-identical task success can differ sharply in policy adherence, a distinction that marginal metrics conceal but the conditional decomposition makes explicit.

On the practical side, we analyzed the annotation allocation problem and demonstrated that retrieval-success annotations are more informative than task-success annotations for estimating policy adherence. This asymmetry has a structural explanation: observing retrieval resolves generator success in three of four joint cells, whereas observing task success resolves only one. We further extended the model to incorporate LLM-as-a-judge annotations as calibrated noisy observations. Our results show that when the judge exhibits high false-positive rates, even thousands of automated annotations yield only marginal gains over a small set of human judgments, underscoring the importance of calibration quality.

More broadly, the framework illustrates how casting evaluation as probabilistic inference enables reasoning about uncertainty, partial observability, and annotation efficiency within a unified model. Natural extensions include continuous quality scales and more complex pipelines incorporating reranking, query reformulation, or iterative retrieval. Future work could also consider a hierarchical model that shares statistical strength across related datasets, retrievers, and generators, rather than treating each RAG configuration independently.

\section*{Limitations}

\paragraph{Binary Model.} The core limitation of this work is the simplified assumption that all judgments are provided on a binary scale (fail vs. success). Especially in Information Retrieval, the use of continuous metrics such as MAP and MRR is common and yields a more fine-grained model.

\paragraph{Deterministic Policy Definition. } Generator success is defined by a single fixed policy (abstain iff retrieval fails, answer correctly otherwise). In practice, reasonable policies may differ; for instance, a system might legitimately attempt a partial answer when retrieval is incomplete, or abstention thresholds may be application-dependent. So far, the framework does not accommodate soft or alternative policy definitions.

\paragraph{Limited Task and Model Diversity.} Due to budget limitations, the experiments cover three datasets (one fact-checking, two QA) and three relatively small open-weight generators (8–12B parameters). 

\paragraph{Simple RAG Pipeline.}  The framework models a minimal retrieve-then-generate pipeline. Current production RAG systems often include additional components such as query reformulation, reranking, chunk filtering, or multi-turn retrieval. Each additional component introduces its own failure modes and dependencies that the current model does not capture. Extending the framework to deeper pipelines would require additional variables and a more complex dependency structure.

\paragraph{Single-Turn Evaluation.} The framework evaluates each query in isolation as a single-turn interaction. It does not account for multi-turn conversational RAG settings, where retrieval and generation decisions depend on dialogue history, and where errors in earlier turns can compound across the conversation.

\paragraph{Abstention Detection.} Abstention is detected through exact matching against the prescribed abstention response. This is appropriate in our constrained generation setting, where answers are restricted to a single entity, number, or fixed label, but would not capture hedged or indirect abstentions in free-form generation. Such settings could instead incorporate a calibrated abstention classifier as a noisy observation model.

\section*{Acknowledgments}

This work was supported by the Swiss National Science Foundation (SNF) within the project "Unified Model for Evaluation of Text Generation Systems (UniVal)" [200020\_219819].

\bibliography{custom}

\appendix

\section{Experimental Setup}\label{sec:experimental_setup}

\subsection{Retrieval}

\noindent Table~\ref{tab:setup_retrieval} summarizes the retrieval configuration.
All retrievers use top-$K = 5$.

\begin{table}[h]
\centering
\small
\begin{tabular}{ll}
\toprule
\multicolumn{2}{l}{\textit{Sparse (BM25)}} \\
\midrule
Tokenizer           & Whitespace \\
Stemmer             & English \\
Stopwords           & English \\
Min.\ DF            & 1 \\
Lowercase           & True \\
Ampersand norm.     & True \\
Special char norm.  & True \\
Acronym norm.       & True \\
Punctuation removal & True \\
\midrule
\multicolumn{2}{l}{\textit{Dense}} \\
\midrule
Embedding model     & \texttt{jinaai/jina-embeddings-v3} \\
Embedding task      & \texttt{text-matching} \\
Retrieval task      & \texttt{retrieval.query} \\
HNSW dim.           & 1024 \\
HNSW $M$            & 32 \\
HNSW metric         & Inner product \\
\midrule
\multicolumn{2}{l}{\textit{Hybrid}} \\
\midrule
RRF $k$             & 1.0 \\
\bottomrule
\end{tabular}
\caption{Retrieval hyperparameters.}
\label{tab:setup_retrieval}
\end{table}

\subsection{Generation}

\begin{table}[h]
\centering
\small
\begin{tabular}{ll}
\toprule
\multicolumn{2}{l}{\textit{Generator}} \\
\midrule
Temperature              & 0.7 \\
Max tokens               & 512 \\
Max concurrent requests  & 15 \\
Thinking                 & None \\
\midrule
\multicolumn{2}{l}{\textit{LLM-as-a-Judge}} \\
\midrule
Model                    & \texttt{gpt-4o-mini-2024-07-18} \\
Max tokens               & 1 \\
Temperature              & 1 \\
Logprobs                 & True \\
Top logprobs             & 20 \\
\bottomrule
\end{tabular}
\caption{Generation and judge hyperparameters.}
\label{tab:setup_generation}
\end{table}

\subsection{Evaluation}
No normalization, article stripping, or case adjustment is applied to
generated answers prior to matching against the ground truth.
Abstention is detected via exact string matching: for FEVER, the label
\texttt{NOT\_ENOUGH\_INFO} is matched; for HotpotQA and NQ, the output
\texttt{I DO NOT KNOW} is matched.

\section{Generation Prompts}\label{sec:prompts}

\subsection{System Prompts}
Depending on the task, a different system prompt is used.

\subsubsection{Fact Checking}
Used for the Fever task.

\begin{quote}
You are a master fact checker.
From given passages and a claim
you can say whether the claim is:
SUPPORTS, REFUTES.

If the passages do not provide a
clear answer you need to say:
"NOT ENOUGH INFO".

Only answer with one of the given labels:
SUPPORTS, REFUTES, NOT ENOUGH INFO.
\end{quote}

\subsubsection{QA}
Used for both the HotpotQA and Natural Question task. \\

\begin{quote}
You are a precise question-answering
assistant. You will be given a question
and a list of retrieved paragraphs
containing the answer. Your response
must be ONLY the answer itself.
A single word, name, entity, or number.
No explanation, no punctuation,
no full sentences, no preamble.
Only use the retrieved paragraphs
to answer the question!
If the answer is not in the retrieved
paragraphs you must say: I DO NOT KNOW

Examples: \\
Q: What nationality is Friedrich Merz? \\
A: German

Q: How many siblings did Marie Curie have? \\
A: Four \\
\end{quote}

\subsubsection{User Prompts}
For all tasks the same user prompt is used.
It consists of a list of paragraphs and the original query.

\begin{verbatim}
{% if retrieved|length > 0 %}
<retrieved>
    {% for p in retrieved %}
    <paragraph
      document_id="{{  p.document_id }}"
      index="{{  p.index }}"
    >
       {{  p.text }}
    </paragraph>
    {% endfor %}
</retrieved>
{% endif %}

<question>
    {{ input }}
</question>
\end{verbatim}

\section{Partial Retrieval Success}\label{sec:partial-success-conditionals}

In Section~\ref{sec:model}, we define retrieval success $R$ as binary and measure downstream abstention and task behavior conditional on this binary outcome. In practice, however, a generation model may be able to produce a correct answer from only partial context~\footnote{This is in part due to relevance annotations not distinguishing which documents are individually sufficient or jointly necessary.}. Here, we add an experiment studying this setting. We define a ternary retrieval outcome $R \in {\text{fail}, \text{partial}, \text{success}}$, where failure means that no relevant documents were retrieved, partial retrieval means that some but not all relevant documents were retrieved, and success means that all relevant documents were retrieved. 

Table~\ref{tab:partial-retrieval} shows the number and proportion of partial retrievals for each dataset and retriever. For \emph{NQ}, the rate of partial retrieval is 0\%, since each query in this dataset has exactly one relevant document. Similarly, the majority of \emph{FEV} queries have exactly one relevant document, while a minority have more than one, leading to a maximum of 453 partial retrievals with the \emph{Hybrid} retriever. For \emph{HQA}, on the other hand, every query has two relevant documents by construction, since \emph{HQA} is designed for multi-hop reasoning. Consequently, its partial retrieval rate reaches up to 56.47

We calculate the conditional decomposition for this ternary setup in Table~\ref{tab:partial-success-conditionals}. We define the conditional probability parameters analogously to Section~\ref{sec:model}:
$\theta_{A,r} = P(A=1 \mid R=r)$ and
$\theta_{T,r} = P(T=1 \mid A=0, R=r)$,
where $r \in {f,p,s}$ corresponds to failure, partial retrieval, and success, respectively. The results are overall consistent with the discussion in Section~\ref{sec:main-res}. In particular, the abstention rate decreases monotonically as retrieval quality increases, while the task success rate increases.

\begin{table}
\centering
\small
\input{tables/partial-retrieval}
\caption{Number of partial retrievals for each dataset and retriever.}\label{tab:partial-retrieval}
\end{table}

\begin{table*}[ht!]
    \centering
    \small
    \input{tables/hotpotqa-partial-success-conditionals}
    \caption{
    Count-based conditional probabilities for HotpotQA treating retrieval outcome as ternary. We show the highest value in {\bf{bold}} and the lowest in \it{italics}.
    }
    \label{tab:partial-success-conditionals}
\end{table*}

\section{LLM-as-a-Judge Prompts}\label{sec:prompts_llmjudge}
The following prompts are used for the LLM-as-a-judge results.

\subsection{System Prompts}
System prompts differ between the two judges.

\subsubsection{Retrieval Judge}
\begin{quote}
Given the following question and documents,
you must analyze the provided documents and
determine whether they are sufficient for
answering the question.
In your evaluation, you should consider
the content of the documents and
how they relate to the provided question.
Output your final verdict by strictly
following this format:
"Yes" if the documents are sufficient and
"No" if the documents provided are
not sufficient. Do not provide any additional
explanation for your decision.
\end{quote}

\subsubsection{Task Judge}
\begin{quote}
Given the following question, documents,
and answer, you must analyze
the provided answer and documents before
determining whether the answer is relevant
for the provided question.
In your evaluation, you should consider
whether the answer addresses all
aspects of the question and
provides only correct information from
the documents for answering the question.
Output your final verdict by strictly
following this format: "Yes" if the answer
is relevant for the given question and
"No" if the answer is not relevant for
the given question. Do not provide any
additional explanation for your decision.
\end{quote}

\subsection{User Prompts}
Both the retrieval and task judge use the same user prompt template.
However the retrieval judge does not see the generated answer.

\begin{verbatim}
<context>
    {% for document in context %}
        <document>
            {{ document }}
        </document>
    {%  endfor %}
</context>

<question>{{ query }}</question>

{% if response|length > 0 %}
    <answer>{{ response }}</answer>
{% endif %}
\end{verbatim}






\section{Extended Sample Allocation Plot}\label{sec:full_allocation}

\begin{figure*}
    \centering
    \includegraphics[width=1\linewidth]{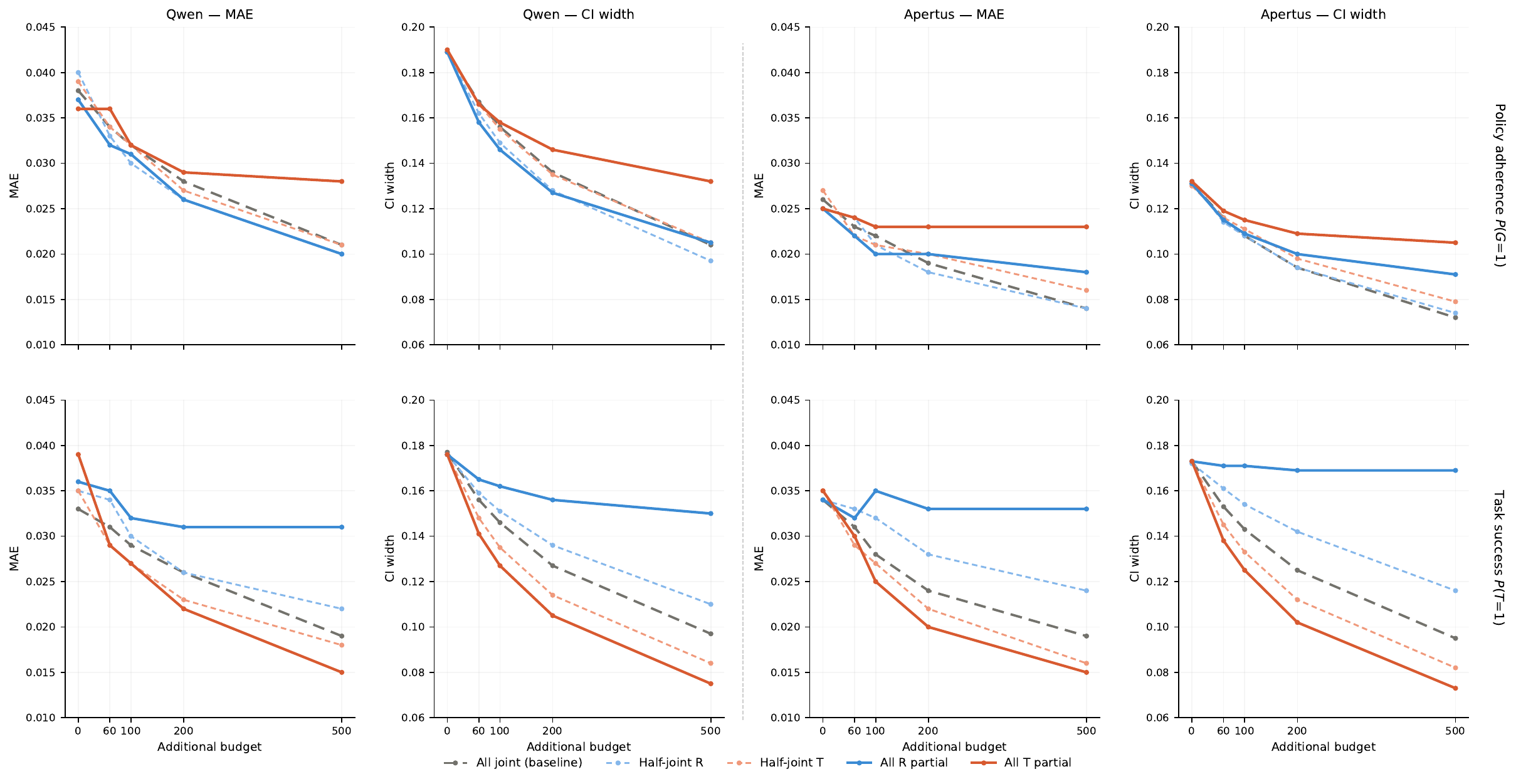}
    \caption{Estimation quality (MAE and 95\% credible interval width) for policy adherence $P(G=1)$ (top) and task success $P(T=1)$ (bottom) under five annotation allocation strategies, for Qwen (left) and Apertus (right). All configurations start with 100 fully annotated base samples and increase the additional budget. Abstention is always observed. Results on HotpotQA with Hybrid retrieval, averaged over 500 subsamples.}
    \label{fig:allocation_full}
\end{figure*}

Figure~\ref{fig:allocation_full} reports both MAE and 95\% credible interval width for policy adherence $P(G{=}1)$ and task success $P(T{=}1)$ across all five annotation strategies, complementing the MAE-only summary in the main text (Figure~\ref{fig:allocation}). The credible interval width closely tracks the MAE ordering: for $P(G{=}1)$, retrieval-focused strategies yield narrower intervals than task-focused ones at every budget level, while for $P(T{=}1)$ the pattern reverses. For instance, at budget 500, the All-R strategy achieves a CI width of 0.105 for Qwen's $P(G{=}1)$ compared to 0.132 for All-T, whereas for $P(T{=}1)$ the All-T strategy reaches 0.075 versus 0.150 for All-R. Coverage remains close to the nominal 95\% level across all strategies and budget levels, indicating that the posterior intervals are well-calibrated regardless of the allocation strategy. The all-joint strategy consistently yields among the narrowest intervals for both quantities, confirming its robustness as a default when the estimation target is unknown in advance.

\section{Information Gain Calculations}
\label{sec:info_gain}
\begin{table}[t]
\centering
\small
\begin{tabular}{llccc}
\toprule
$R$ & $A$ & $G$ & Qwen & Apertus \\
\midrule
0 & 1 & $1$ & 0.367 & 0.031 \\
0 & 0 & $0$ & 0.426 & 0.762 \\
1 & 1 & $0$ & 0.007 & 0.001 \\
1 & 0 & $T$ & 0.200 & 0.206 \\
\bottomrule
\end{tabular}
\caption{Joint probabilities $P(R,A)$ for each cell of the $(R,A)$ contingency table, with the resulting value of $G$. In three of four cells, $G$ is fully determined by $(R,A)$ alone. Only the cell $(R{=}1, A{=}0)$ leaves $G$ unresolved, where $G=T$. Values shown for Qwen and Apertus on HotpotQA with Hybrid retrieval.}
\label{tab:ra_cells}
\end{table}

We now analyze why certain annotation strategies are more effective than others for estimating policy adherence $P(G{=}1)$. Since abstention is always observed via string matching, we condition on $A$ throughout and ask: how much does additionally observing $R$, $T$, or both reduce our uncertainty about $G$?

\paragraph{Joint Probability Structure.}
Recall that $G$ is defined as:
\begin{align}
\small
G = (R{=}0 \land A{=}1) \lor (R{=}1 \land A{=}0 \land T{=}1). \nonumber
\end{align}
Table~\ref{tab:ra_cells} shows the four $(R, A)$ cells and whether $G$ is determined. In three cells, $G$ follows directly from $R$ and $A$---these correspond to cases where either retrieval or abstention went wrong. Only when both retrieval and abstention are correct, i.e., $(R{=}1, A{=}0)$, does $G$ depend on $T$. This asymmetry is the structural basis for the annotation efficiency differences we observe.

\paragraph{Base Uncertainty.}
The uncertainty about $G$ given only $A$ is:
\begin{align}
H(G \mid A) = \sum_{a} P(A{=}a) \cdot H(G \mid A{=}a).
\end{align}
When $A{=}1$ (the generator abstained), $G$ depends only on whether the abstention was justified, i.e., whether $R{=}0$. Since $R$ is unobserved:
\begin{align}
H(G \mid A{=}1) = h\!\left(\frac{P(A{=}1, R{=}0)}{P(A{=}1)}\right),
\end{align}
where $h(\cdot)$ denotes the binary entropy function. For Qwen, $P(R{=}0 \mid A{=}1) = 0.367 / 0.374 = 0.981$, yielding $h(0.981) = 0.134$; for Apertus, $0.031 / 0.032 = 0.969$, yielding $h(0.969) = 0.196$.

When $A{=}0$ (the generator answered), $G{=}1$ requires both $R{=}1$ and $T{=}1$. With neither observed:
\begin{align}
\small
H(G \mid A{=}0) &= h\!\bigl(P(G{=}1 \mid A{=}0)\bigr)
\end{align}

where 
\begin{align}
\small
P(G{=}1 \mid A{=}0) = \frac{P(R{=}1)\, P(A{=}0 \mid R{=}1)\, \theta_{T^+}}{P(A{=}0)}
\end{align}
Remember $\theta_{T^+} = P(T=1 \mid R=1, A=0)$.

For Qwen, $P(G{=}1 \mid A{=}0) = 0.216$, giving $h(0.216) = 0.760$; for Apertus, $P(G{=}1 \mid A{=}0) = 0.110$, giving $h(0.110) = 0.500$.

Combining both cases:
\begin{align}
H(G \mid A) &= P(A{=}1) \cdot h_{A=1} + P(A{=}0) \cdot h_{A=0},
\end{align}
yielding $0.526$ for Qwen and $0.490$ for Apertus. Despite very different abstention behaviors, both models exhibit similar base uncertainty---though from different sources: for Qwen, a balanced split between $G{=}1$ and $G{=}0$ in the $A{=}0$ cell; for Apertus, the sheer mass of the $A{=}0$ cell (96.8\% of samples) compensates for its lower per-sample entropy.

\paragraph{All-Joint Strategy.}
Observing both $R$ and $T$ resolves $G$ completely, since $G$ is a deterministic function of $(R, A, T)$:
\begin{align}
I(G; R, T \mid A) &= H(G \mid A) - \underbrace{H(G \mid R, A, T)}_{=\,0} \nonumber \\
&= H(G \mid A).
\end{align}
This represents the maximum achievable information gain per sample.

\paragraph{All-R Strategy.}
Observing only $R$ yields:
\begin{align}
I(G; R \mid A) = H(G \mid A) - H(G \mid R, A).
\end{align}
From Table~\ref{tab:ra_cells}, three of four $(R, A)$ cells resolve $G$ deterministically. Only $(R{=}1, A{=}0)$ leaves $G$ unresolved, where $G = T$. The residual uncertainty is therefore:
\begin{align}
H(G \mid R, A) = P(R{=}1, A{=}0) \cdot h(\theta_{T^+}).
\end{align}
For Qwen: $0.200 \times h(0.678) = 0.181$; for Apertus: $0.206 \times h(0.519) = 0.206$. The resulting gains are $I(G; R \mid A) = 0.345$ (Qwen) and $0.284$ (Apertus). Observing $R$ eliminates the majority of the uncertainty about $G$.

\paragraph{All-T Strategy.}
Observing only $T$ yields:
\begin{align}
I(G; T \mid A) = H(G \mid A) - H(G \mid A, T).
\end{align}
Of the three non-zero $(A, T)$ cells, only $(A{=}0, T{=}0)$ resolves $G$ deterministically ($G{=}0$, since both branches of the definition fail). The remaining two cells require $R$:
\begin{itemize}[nosep,leftmargin=*]
\item $(A{=}1, T{=}0)$: the generator abstained, but $G$ depends on whether the abstention was justified, i.e., $H = h(P(R{=}0 \mid A{=}1))$.
\item $(A{=}0, T{=}1)$: the generator answered correctly, but $G$ depends on whether the answer was supported, i.e., $H = h(P(R{=}1 \mid A{=}0, T{=}1))$.
\end{itemize}
The residual uncertainty is:
\begin{align}
H(G \mid A, T) &= P(A{=}1) \cdot h\!\left(P(R{=}0 \mid A{=}1)\right) \nonumber \\
&\quad + P(A{=}0, T{=}1) \nonumber \\
&\quad \cdot h\!\left(P(R{=}1 \mid A{=}0, T{=}1)\right),
\end{align}
where $P(R{=}1 \mid A{=}0, T{=}1)$ is obtained via Bayes' rule:
\begin{align}
\frac{P(R{=}1)\, P(A{=}0 \mid R{=}1)\, \theta_{T^+}}{P(A{=}0, T{=}1)}.
\end{align}
For Qwen: $P(R{=}1 \mid A{=}0, T{=}1) = 0.411$, yielding $H(G \mid A, T) = 0.372$ and $I(G; T \mid A) = 0.154$. For Apertus: $P(R{=}1 \mid A{=}0, T{=}1) = 0.352$, yielding $H(G \mid A, T) = 0.290$ and $I(G; T \mid A) = 0.200$.

\paragraph{Half-Joint-R Strategy.}
Here, half the additional budget is allocated to joint samples and the other half to retrieval-only samples. Each joint sample contributes $I(G; R, T \mid A) = H(G \mid A)$, while each R-partial sample contributes $I(G; R \mid A)$. The expected per-sample gain is the average:
\begin{align}
\tfrac{1}{2}\, H(G \mid A) + \tfrac{1}{2}\, I(G; R \mid A),
\end{align}
yielding $0.436$ for Qwen and $0.387$ for Apertus. Since the R-partial component already captures most of the uncertainty (three of four cells resolved), the loss relative to the all-joint strategy is modest.

\paragraph{Half-Joint-T Strategy.}
Analogously, half the budget goes to joint samples and half to task-only samples. The expected per-sample gain is:
\begin{align}
\tfrac{1}{2}\, H(G \mid A) + \tfrac{1}{2}\, I(G; T \mid A),
\end{align}
yielding $0.340$ for Qwen and $0.345$ for Apertus. Despite receiving the same number of joint samples as Half-Joint-R, this strategy is less effective because the T-partial component contributes less information about $G$---it leaves two cells unresolved rather than one.

\paragraph{Summary.}
Table~\ref{tab:info_gain_G} summarizes the information gains. The structural asymmetry is clear: observing $R$ leaves one ambiguous cell for $G$, while observing $T$ leaves two. This directly explains why retrieval-focused strategies outperform task-focused strategies for estimating policy adherence. The mixed strategies interpolate linearly between these extremes.

\paragraph{Discussion.}
The information-theoretic ranking in Table~\ref{tab:info_gain_G} correctly predicts the relative ordering of partial strategies: All-R consistently outperforms All-T for estimating $P(G{=}1)$, and the half-joint strategies fall between the corresponding extremes. However, the predicted ranking does not perfectly match the empirical results at all budget levels. For Apertus, the empirical MAE ordering at budget 500 follows the theoretical ranking closely: All-Joint $\approx$ Half-Joint-R $<$ Half-Joint-T $<$ All-R $<$ All-T. For Qwen, however, All-R and Half-Joint-R slightly outperform All-Joint despite having lower per-sample information gain.

This discrepancy arises because the information-theoretic analysis treats each sample independently, whereas the Bayesian model shares information across cells through its parameterization. The only cell left unresolved by $(R, A)$ observations is $(R{=}1, A{=}0)$, where $G = T$. The 100 base joint samples provide direct observations of $T$ in this cell, allowing the model to estimate $\theta_{T^+}$. Once this parameter is sufficiently well-estimated, additional joint samples yield diminishing returns---they annotate $T$ in cells where $G$ is already resolved by $(R, A)$ alone. R-partial samples, by contrast, contribute exclusively to the three resolved cells, where each observation provides a clean binary label for $G$.

For Apertus, $\theta_{T^+} = 0.519$ with near-maximal entropy ($h(0.519) = 0.999$), making it intrinsically harder to estimate. The 100 base samples are insufficient to pin it down, and additional joint samples that directly observe $T$ in the ambiguous cell remain valuable. For Qwen, $\theta_{T^+} = 0.678$ with lower entropy ($h(0.678) = 0.904$), allowing the base samples to estimate it more effectively and reducing the marginal value of additional joint annotations.

In summary, the information-theoretic analysis reliably predicts \emph{which} partial observation is more informative for a given target, while the question of whether partial annotations can fully \emph{substitute} for joint annotations depends on how well the base samples constrain the parameters in the ambiguous cells.

\begin{table}[t]
\centering
\small
\begin{tabular}{lcc}
\toprule
Strategy & Qwen & Apertus \\
\midrule
All-Joint & 0.526 & 0.490 \\
Half-Joint-R & 0.436 & 0.387 \\
All-R & 0.345 & 0.284 \\
Half-Joint-T & 0.340 & 0.345 \\
All-T & 0.154 & 0.200 \\
\bottomrule
\end{tabular}
\caption{Information gain $I(G; \cdot \mid A)$ per sample for each annotation strategy, computed from the conditional probabilities in Table~\ref{tab:rag-conditionals}. Higher values indicate more informative observations for estimating $P(G{=}1)$. The ranking is consistent with the empirical MAE ordering in Figure~\ref{fig:allocation}.}
\label{tab:info_gain_G}
\end{table}

\section{AI Assistants}
AI assistants (Claude, Anthropic) were used for code development, data analysis, and iterative drafting of manuscript text. All outputs were reviewed and validated by the authors.

\end{document}

%% file: tables/naive_all_marginal_point_estimates.tex
\begin{tabular}{lllcccc}
\toprule
DS & Gen. & Ret. & $P(R{=}1)$ & $P(A{=}1)$ & $P(T{=}1)$ & $P(G{=}1)$ \\
\midrule
\multirow{9}{*}{FEV}
 & Apt & \multirow{3}{*}{D} & \multirow{3}{*}{0.603} & 0.135 & 0.753 & 0.608 \\
 & Gem &  &  & 0.180 & 0.784 & 0.736 \\
 & Qwn &  &  & 0.199 & 0.751 & \bf{0.744} \\
\cmidrule(lr){2-7}
 & Apt & \multirow{3}{*}{H} & \multirow{3}{*}{\bf{0.655}} & \it{0.089} & 0.801 & 0.614 \\
 & Gem &  &  & 0.096 & \bf{0.864} & 0.701 \\
 & Qwn &  &  & 0.112 & 0.838 & 0.703 \\
\cmidrule(lr){2-7}
 & Apt & \multirow{3}{*}{S} & \multirow{3}{*}{0.456} & 0.124 & 0.759 & \it{0.504} \\
 & Gem &  &  & 0.221 & 0.744 & 0.644 \\
 & Qwn &  &  & \bf{0.246} & \it{0.708} & 0.657 \\
\midrule
\multirow{9}{*}{HQA}
 & Apt & \multirow{3}{*}{D} & \multirow{3}{*}{0.151} & 0.038 & 0.260 & \it{0.110} \\
 & Gem &  &  & 0.311 & \it{0.253} & 0.391 \\
 & Qwn &  &  & \bf{0.489} & 0.260 & \bf{0.574} \\
\cmidrule(lr){2-7}
 & Apt & \multirow{3}{*}{H} & \multirow{3}{*}{0.207} & \it{0.032} & 0.304 & 0.138 \\
 & Gem &  &  & 0.218 & 0.315 & 0.337 \\
 & Qwn &  &  & 0.375 & \bf{0.329} & 0.503 \\
\cmidrule(lr){2-7}
 & Apt & \multirow{3}{*}{S} & \multirow{3}{*}{\bf{0.218}} & 0.033 & 0.318 & 0.152 \\
 & Gem &  &  & 0.226 & 0.318 & 0.360 \\
 & Qwn &  &  & 0.391 & 0.326 & 0.535 \\
\midrule
\multirow{9}{*}{NQ}
 & Apt & \multirow{3}{*}{D} & \multirow{3}{*}{0.347} & 0.028 & 0.239 & 0.164 \\
 & Gem &  &  & 0.256 & 0.242 & 0.414 \\
 & Qwn &  &  & 0.355 & 0.243 & 0.496 \\
\cmidrule(lr){2-7}
 & Apt & \multirow{3}{*}{H} & \multirow{3}{*}{\bf{0.348}} & \it{0.024} & \bf{0.267} & 0.163 \\
 & Gem &  &  & 0.198 & 0.267 & 0.359 \\
 & Qwn &  &  & 0.307 & 0.266 & 0.451 \\
\cmidrule(lr){2-7}
 & Apt & \multirow{3}{*}{S} & \multirow{3}{*}{0.244} & 0.039 & 0.233 & \it{0.136} \\
 & Gem &  &  & 0.300 & \it{0.220} & 0.412 \\
 & Qwn &  &  & \bf{0.434} & 0.215 & \bf{0.531} \\
\bottomrule
\end{tabular}

%% file: tables/conditional_point_estimates_compressed.tex
\begin{tabular}{lllcccc}
\toprule
DS & Gen. & Ret. & $\theta_{A^-}$ & $\theta_{A^+}$ & $\theta_{T^-}$ & $\theta_{T^+}$ \\
\midrule
\multirow{9}{*}{FEV}
 & Apt & \multirow{3}{*}{D} & 0.245 & \bf{0.063} & \it{0.810} & 0.903 \\
 & Gem &  & 0.422 & 0.021 & 0.941 & 0.962 \\
 & Qwn &  & \bf{0.462} & 0.027 & 0.894 & 0.954 \\
\cmidrule(lr){2-7}
 & Apt & \multirow{3}{*}{H} & \it{0.158} & 0.053 & 0.831 & \it{0.901} \\
 & Gem &  & 0.242 & 0.019 & 0.940 & 0.961 \\
 & Qwn &  & 0.274 & 0.027 & 0.913 & 0.955 \\
\cmidrule(lr){2-7}
 & Apt & \multirow{3}{*}{S} & 0.197 & 0.038 & 0.828 & 0.906 \\
 & Gem &  & 0.391 & \it{0.018} & \bf{0.942} & \bf{0.963} \\
 & Qwn &  & 0.428 & 0.028 & 0.916 & 0.956 \\
\midrule
\multirow{9}{*}{HQA}
 & Apt & \multirow{3}{*}{D} & 0.044 & 0.010 & \it{0.229} & \it{0.492} \\
 & Gem &  & 0.361 & 0.026 & 0.313 & 0.570 \\
 & Qwn &  & \bf{0.568} & \bf{0.044} & \bf{0.458} & 0.638 \\
\cmidrule(lr){2-7}
 & Apt & \multirow{3}{*}{H} & \it{0.039} & \it{0.005} & 0.258 & 0.519 \\
 & Gem &  & 0.270 & 0.021 & 0.331 & 0.608 \\
 & Qwn &  & 0.463 & 0.036 & 0.456 & 0.678 \\
\cmidrule(lr){2-7}
 & Apt & \multirow{3}{*}{S} & 0.041 & 0.006 & 0.264 & 0.552 \\
 & Gem &  & 0.286 & 0.011 & 0.326 & 0.632 \\
 & Qwn &  & 0.493 & 0.028 & 0.446 & \bf{0.704} \\
\midrule
\multirow{9}{*}{NQ}
 & Apt & \multirow{3}{*}{D} & 0.039 & 0.007 & \it{0.161} & \it{0.402} \\
 & Gem &  & 0.382 & 0.021 & 0.192 & 0.485 \\
 & Qwn &  & 0.513 & 0.060 & 0.256 & 0.494 \\
\cmidrule(lr){2-7}
 & Apt & \multirow{3}{*}{H} & \it{0.034} & \it{0.006} & 0.179 & 0.408 \\
 & Gem &  & 0.290 & 0.024 & 0.210 & 0.499 \\
 & Qwn &  & 0.439 & 0.061 & 0.277 & 0.503 \\
\cmidrule(lr){2-7}
 & Apt & \multirow{3}{*}{S} & 0.050 & 0.009 & 0.187 & 0.407 \\
 & Gem &  & 0.387 & 0.032 & 0.216 & 0.505 \\
 & Qwn &  & \bf{0.550} & \bf{0.077} & \bf{0.293} & \bf{0.513} \\
\bottomrule
\end{tabular}

%% file: tables/llm_judge_metrics_table.tex
\begin{tabular}{rcccc}
\toprule
 & \multicolumn{2}{c}{P(G=1)} & \multicolumn{2}{c}{P(T=1)} \\
$n_{\mathrm{add}}$ & MAE & CI W. & MAE & CI W. \\
\midrule
0 & 0.0174 & 0.0941 & 0.0232 & 0.1246 \\
500 & 0.0166 & 0.0838 & 0.0243 & 0.1176 \\
5000 & 0.0170 & 0.0802 & 0.0230 & 0.1150 \\
\bottomrule
\end{tabular}

%% file: tables/partial-retrieval.tex
\begin{tabular}{llr}
\toprule
Dataset & Retriever & Partial retrievals, $n$ (\%) \\
\midrule
\multirow{3}{*}{FEVER}
 & Dense  & 419 (4.19\%) \\
 & Hybrid & 453 (4.53\%) \\
 & Sparse & 300 (3.00\%) \\
\midrule
\multirow{3}{*}{HotpotQA}
 & Dense  & 4,724 (47.24\%) \\
 & Hybrid & 5,647 (56.47\%) \\
 & Sparse & 5,233 (52.33\%) \\
\midrule
\multirow{3}{*}{NQ}
 & Dense  & 0 (0.00\%) \\
 & Hybrid & 0 (0.00\%) \\
 & Sparse & 0 (0.00\%) \\
\bottomrule
\end{tabular}

%% file: tables/hotpotqa-partial-success-conditionals.tex
\begin{tabular}{lllcccccc}
\toprule
DS & Gen. & Ret.
& $\theta_{A,f}$
& $\theta_{A,p}$
& $\theta_{A,s}$
& $\theta_{T,f}$
& $\theta_{T,p}$
& $\theta_{T,s}$ \\
\midrule
\multirow{9}{*}{HQA}
 & Apt & \multirow{3}{*}{D}
 & 0.064 & \it{0.027} & 0.009 & \it{0.120} & 0.313 & \it{0.492} \\
 & Gem & 
 & 0.592 & 0.177 & 0.025 & 0.143 & 0.379 & 0.570 \\
 & Qwn & 
 & \bf{0.812} & \bf{0.373} & \bf{0.044} & \bf{0.272} & \bf{0.502} & 0.638 \\
\cmidrule(lr){2-9}
 & Apt & \multirow{3}{*}{H}
 & \it{0.058} & 0.031 & \it{0.005} & 0.127 & \it{0.310} & 0.519 \\
 & Gem & 
 & 0.518 & 0.169 & 0.020 & 0.158 & 0.372 & 0.608 \\
 & Qwn & 
 & 0.723 & 0.358 & 0.036 & 0.242 & 0.493 & 0.679 \\
\cmidrule(lr){2-9}
 & Apt & \multirow{3}{*}{S}
 & 0.068 & 0.028 & \it{0.005} & 0.146 & 0.321 & 0.553 \\
 & Gem & 
 & 0.525 & 0.168 & 0.011 & 0.155 & 0.374 & 0.633 \\
 & Qwn & 
 & 0.744 & 0.369 & 0.027 & 0.208 & 0.494 & \bf{0.705} \\
\bottomrule
\end{tabular}